\documentclass{sharpa}
\usepackage{amsmath,amssymb,booktabs,array,tabularx,float}
\usepackage{url}
\setcitestyle{numbers,sort&compress}
\title{DexForesight: Grounding Dexterous Policies with Implicit Spatial Foresight}
\author[1,2]{Zhongxi Chen}
\author[1,2]{Yi Yang}
\author[4]{Xinyi Chen}
\author[1,2]{Mingzhe Ye}
\author[1,3]{Yaping Li}
\author[1,2]{Zhihong Liu}
\author[1,2]{Congsheng Xu}
\author[1]{Haonan Wang}
\author[2]{Zhijie Deng}
\affiliation[1]{Sharpa Robotics}
\affiliation[2]{SJTU-Sharpa Joint Lab for Embodied AI, SJTU}
\affiliation[3]{CUHK}
\affiliation[4]{FDU}
\abstract{

    Dexterous manipulation requires rich spatial foresight over both the surrounding scene and the robot hand. Existing world-action models often acquire such predictive knowledge through explicit future prediction. We introduce DexForesight, a framework that learns spatial foresight implicitly by injecting dynamics-aware world representations into the latent space of a dexterous policy. During training, DexForesight learns an action-conditioned spatial world model from offline scene depth, kinematic hand geometry, and demonstration trajectories. The world model predicts 3D hand-point motion and visual feature changes, capturing both articulated hand dynamics and scene changes under demonstrated actions, including the motion of fingers occluded from the cameras. We then freeze the world model and align policy queries with its predictive world queries, transferring the learned spatial dynamics into policy latents and using them to condition action generation. At deployment, only the policy is retained, operating on RGB observations, language, and robot state without depth input or explicit future prediction. On 11 DexJoCo simulation tasks, DexForesight improves the overall success rate of the base policy from 49.4\% to 60.5\%. On three real-robot dexterous manipulation tasks, DexForesight achieves 58.3\% success, compared with 50.0\% for the strongest baseline.
    {\small\noindent
    \textbf{Project page:}\enspace
    \href{https://dexforesight.github.io/}{https://dexforesight.github.io/}\par}

}
\metadata[Correspondence]{\url{zhijied@sjtu.edu.cn}}
\renewcommand{\titlelogo}{sharpa-logo.png}
\renewcommand{\titlelogowidth}{2.8cm}
\renewcommand{\titlelogoraise}{-16.7pt}
\renewcommand{\headerlogo}[1]{\ifnum\value{page}=1 \smash{\raisebox{-16.7pt}{\includegraphics[width=2.8cm]{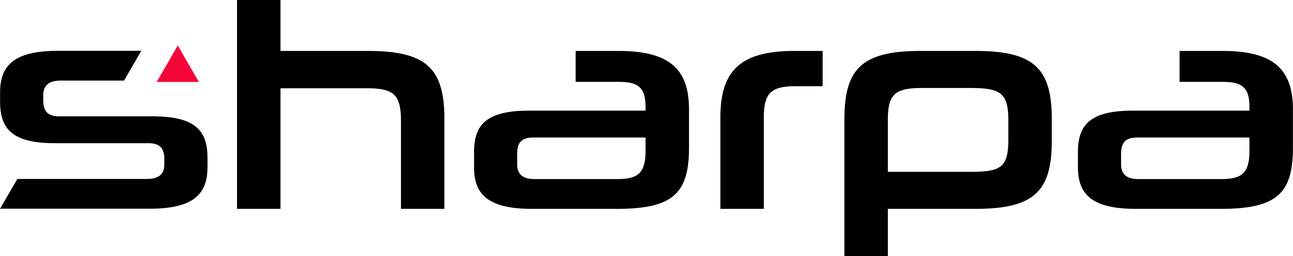}}}\fi}
\hypersetup{pdftitle={DexForesight: Grounding Dexterous Policies with Implicit Spatial Foresight},pdfauthor={Zhongxi Chen, Yi Yang, Xinyi Chen, Mingzhe Ye, Yaping Li, Zhihong Liu, Congsheng Xu, Haonan Wang, Zhijie Deng},pdfkeywords={dexterous manipulation, spatial foresight, world model, robot learning}}
\newcommand{\CA}{\operatorname{CA}}

\begin{document}
\maketitle
\section{Introduction}
\label{sec:introduction}
Controlling a dexterous hand requires anticipating how multiple fingers move relative to an object and the surrounding scene. During a grasp, fingers must move around object surfaces, often while part of the hand is hidden from the cameras. Recognizing the current scene alone is therefore insufficient: the policy also needs information about how actions change the hand--scene configuration. We refer to this predictive understanding as spatial foresight, encompassing both articulated hand motion and changes in the surrounding scene.

World action models (WAMs) couple future prediction with action generation, offering a natural route to learning such foresight~\citep{yuan2026fast,yang2026wla}. Applying them to dexterous manipulation, however, raises two challenges. First, the spatial information used for prediction must capture more than visible scene surfaces. Even accurate scene depth leaves occluded fingers unobserved, whereas the robot's kinematic model and joint states provide hand-link positions independently of camera visibility. These complementary sources can ground predictions in both scene and hand geometry. Yet requiring estimated depth at deployment~\citep{spatialvla,geovla} introduces dependence on an online reconstruction pipeline: depth estimation adds computation, and reliable depth videos require temporal consistency as the hand and objects move~\citep{li2023temporally}. The challenge is to exploit this spatial information for learning without making reconstruction a prerequisite for control.



\begin{figure}[t]
\centering
\includegraphics[width=.90\linewidth]{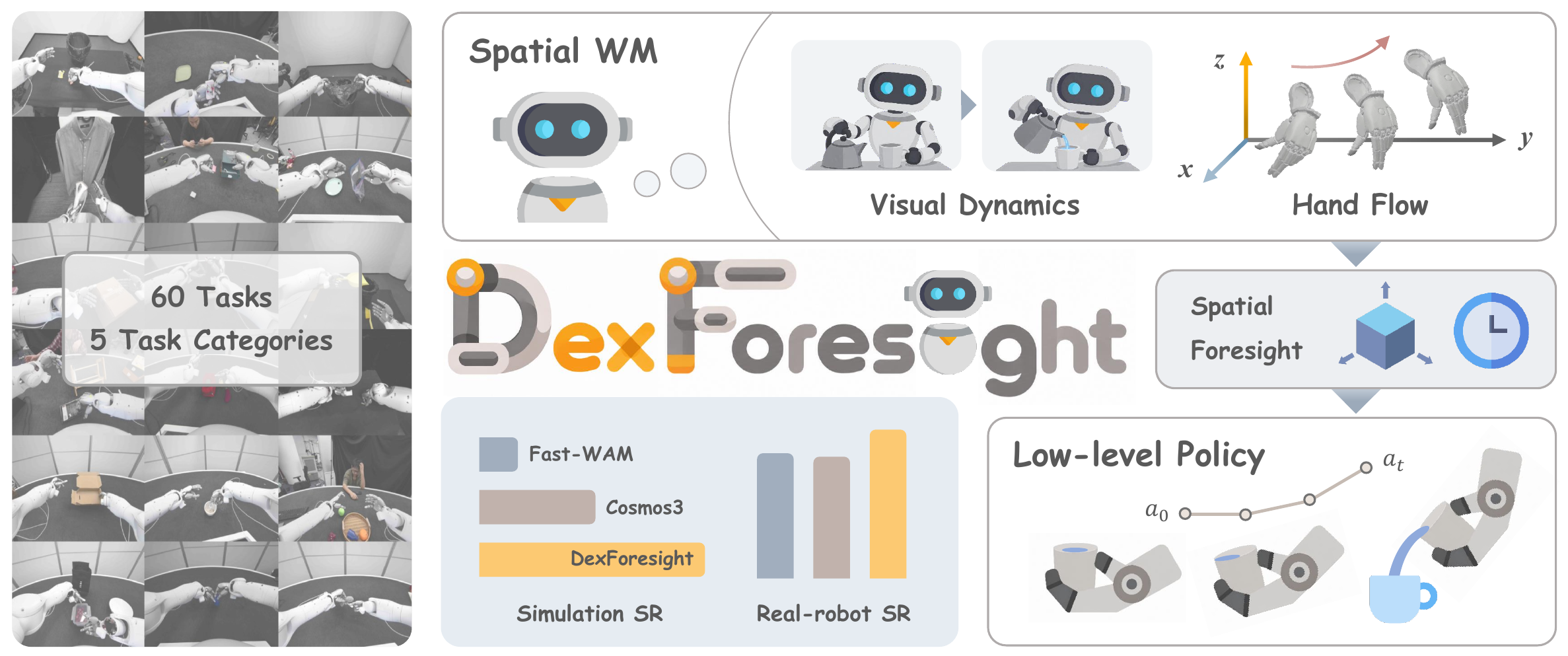}
\caption{\textbf{Learning and transferring spatial foresight.} Demonstrations spanning 60 tasks and five spatial categories train an action-conditioned spatial world model to predict visual dynamics and 3D hand flow. The learned spatial foresight is transferred into a low-level policy, enabling action generation without world-model execution or future decoding at deployment. DexForesight outperforms prior world-action models in both simulation and real-robot evaluations.}
\label{fig:teaser}
\end{figure}

Second, predictive knowledge need not be used through explicit future generation. In WAM designs that imagine the future during inference, the model generates future images, motion, or spatial states alongside, or before, its actions~\citep{videovla}. These predictions then guide action selection or planning~\citep{pointworld}, making future generation an explicit part of the deployed decision process and adding computation at every control step. For dexterous control, however, the key information is how actions change the hand--scene configuration; reconstructing full future observations may additionally require appearance details, such as texture or background, with limited relevance to action selection~\citep{flare}. Can a policy instead acquire the predictive features needed for spatial foresight during training, without decoding any future outputs at deployment?

These challenges motivate transferring spatial foresight into the policy's latent representations. We introduce \textbf{DexForesight: Grounding Dexterous Policies with Implicit Spatial Foresight} (Figure~\ref{fig:teaser}), which decouples an action-conditioned spatial world model from the policy. The action-conditioned world model learns hand and scene dynamics from training-time signals: scene depth estimated offline from stereo demonstrations with StereoAnyVideo~\citep{jing2025stereo}, kinematic hand geometry, and recorded future states. Given demonstrated controls, it jointly predicts displacements of fixed points on the hand, which cover occluded fingers through forward kinematics, and changes in visual features of the observed scene. Both objectives train a shared set of world queries, while depth conditions the backbone on current scene geometry through cross-layer fusion, so no future depth targets are required. 

We then freeze the world model and transfer its predictive features into the policy through query alignment (Figure~\ref{fig:dexforesight}): policy queries read hidden layers of the VLM backbone, match the frozen world queries, and condition the action expert, linking predictive supervision to action learning within the same policy. At deployment, the policy operates only on current RGB observations, language, and robot state, with no world-model execution, online depth reconstruction, or explicit future generation. DexForesight thus acquires spatial foresight of hand and scene dynamics through action-conditioned prediction during training and carries it into dexterous action generation through the policy's own latent features.

Our contributions are:

\begin{itemize}
    \setlength{\itemsep}{4pt}
    \setlength{\parsep}{0pt}
    \item  We introduce \textbf{DexForesight}, which aligns dexterous policy latents with predictive representations from a frozen action-conditioned spatial world model and uses the aligned features to condition action generation, enabling implicit spatial foresight without world-model execution, online depth reconstruction, or explicit future generation at deployment. 
    \item We develop an action-conditioned spatial world model using approximately 370 hours of demonstrations across 60 tasks and five categories. Combining offline scene depth with kinematic hand geometry, it jointly predicts 3D hand motion and visual dynamics, providing supervision beyond visible RGB observations.

    \item We evaluate DexForesight on 11 DexJoCo simulation tasks~\citep{dexjoco} and three real-robot tasks: Close Box, Insert Battery, and Screw On Lid, testing hand--object geometry, alignment under object occlusion, and articulated motion under finger occlusion, respectively. DexForesight achieves 60.5\% and 58.3\% success in simulation and real-robot evaluations, surpassing the strongest respective baselines by 9.6 and 8.3 percentage points.
\end{itemize}


\section{Related Work}

\subsection{Future Modeling for Robot Policies}

World-action models couple future prediction with policy learning, either by jointly modeling future observations and actions~\citep{uwm,videovla,lingbotva} or by incorporating predictive supervision during training without requiring future generation at deployment~\citep{mmact, yuan2026fast,beingh07,flare}.
Beyond image prediction, EgoWAM~\citep{egowam} explores 3D motion flow as a predictive target for human--robot co-training.
These approaches motivate learning structured predictive representations, while dexterous manipulation calls for supervision that captures the coordinated motion of articulated fingers, including those hidden from the cameras.
DexForesight addresses this requirement by combining scene geometry with kinematic hand representations and jointly predicting action-conditioned 3D hand motion and visual changes.
Persistent hand-point targets preserve correspondence across articulated motion and supervise occluded fingers through forward kinematics.
The resulting predictive representations are transferred into policy latents, without requiring world-model execution or future decoding at deployment.

\subsection{Spatially Grounded Policies}

Spatially grounded robot policies introduce geometric structure either directly into the policy~\citep{spatialvla,depthvla,geovla} or through representation-level supervision.
ReconVLA~\citep{reconvla} uses auxiliary reconstruction to improve spatial perception, while Spatial Forcing~\citep{spatialforcing} aligns intermediate VLA features with representations from pretrained 3D foundation models, enabling implicit spatial grounding without explicit 3D inputs at deployment.
These approaches primarily improve representations of scene geometry, rather than explicitly modeling how an articulated dexterous hand evolves under control.
DexForesight instead targets \emph{action-conditioned spatial dynamics}: it combines offline scene depth with robot kinematics and predicts future 3D displacements of persistent hand points together with visual changes.
This embodiment-aware supervision captures hand--scene evolution beyond visible image geometry, including occluded finger motion, while the deployed policy still requires only RGB, language, and robot state.
\section{Method}
\label{sec:method}

\begin{figure}[t]
\centering
\includegraphics[width=.90\linewidth]{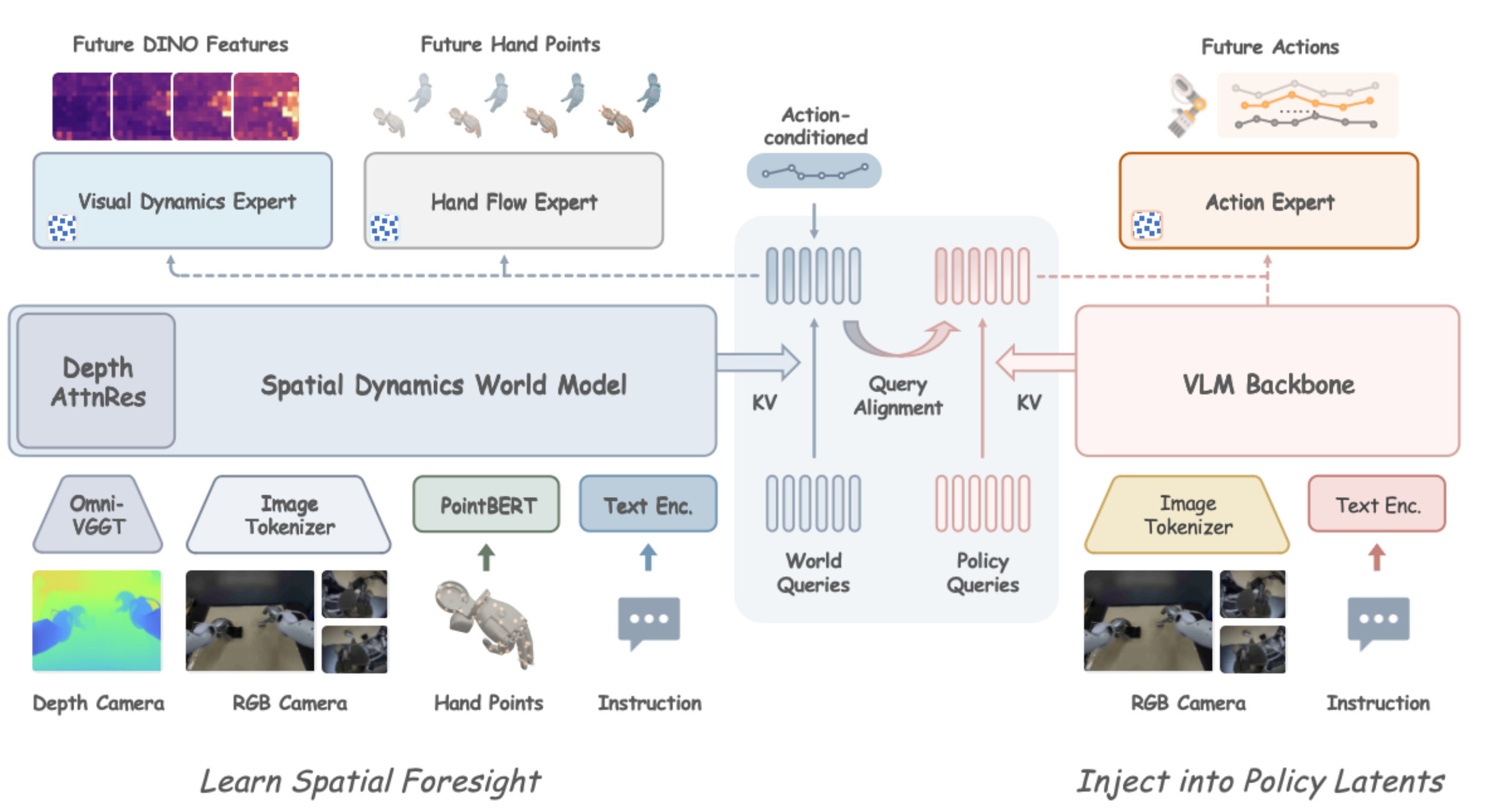}
\caption{\textbf{DexForesight architecture and query alignment.} The spatial world model learns action-conditioned hand motion and visual changes from RGB, offline scene depth, hand geometry, and language. Demonstrated controls condition the world queries through the query adapter and are also supplied to both prediction experts. Queries from the frozen world model supervise policy queries, which condition the action expert.}
\label{fig:dexforesight}
\end{figure}

DexForesight uses offline scene depth and hand geometry computed from recorded robot states to train a spatial world model. Given demonstrated actions, the model predicts how points on the hand move and how image features change. We then freeze the world model and align policy queries with its query features. The policy uses these queries to generate actions from current RGB, language, and robot state (Figure~\ref{fig:dexforesight}), without online depth reconstruction or future decoding.

\paragraph{Overview.}
At time $t$, the policy receives multi-view RGB images, a language instruction, and robot state, denoted by $o_t=(I_t,\ell,s_t)$, and outputs an action chunk $A_t$. The world model additionally receives depth and a short history of hand geometry and robot state; we denote its observation inputs by $c_t$. During training, it also receives a demonstrated control sequence $U_t$ aligned with the future timestamps selected for prediction. The two models produce query features as follows:
\begin{equation}
Z_t^{\mathrm w}=f_\phi(c_t,U_t),
\qquad
Z_t^\pi=q_\theta\!\left(\mathcal M_\theta(o_t)\right).
\label{eq:overview}
\end{equation}
Here $f_\phi$ includes the world-model backbone, query reader, and action-conditioned query adapter. The policy query reader $q_\theta$ attends to selected VLM hidden layers, collected in $\mathcal M_\theta$. Both outputs contain $N$ query tokens. $U_t$ is a training-only dynamics-conditioning variable that resolves which control-dependent future is modeled; it is never an input to the policy. Only the policy runs at deployment.

\subsection{Learning Spatial Foresight}
\label{sec:spatial}

The spatial world model in Figure~\ref{fig:dexforesight} encodes scene observations and hand geometry in a backbone initialized from RynnBrain~\citep{rynnbrain}. Depth enters through intermediate fusion blocks. World queries read the resulting hidden states and receive demonstrated controls through an action-conditioned adapter; these queries condition the Hand Flow Expert and the Visual Dynamics Expert.

\paragraph{Hand geometry.}
We sample fixed points on the hand's URDF links and transform them into the robot base frame through forward kinematics using recorded robot states. Each point retains its link assignment and local coordinates across frames, preserving correspondence as the hand articulates. The resulting point clouds specify the positions of hand surfaces, including those occluded in the camera views. A Point-BERT~\citep{pointbert} encoder encodes the current point cloud and a short history of hand motion. These hand features enter the backbone alongside RGB, language, and robot-state history. Current and historical hand geometry provide encoder inputs, while future hand-point displacements provide the hand-flow prediction targets.

\paragraph{Depth fusion.}
For real-robot demonstrations, we estimate scene depth offline from recorded stereo RGB sequences using StereoAnyVideo. For simulation, we render depth directly from MuJoCo rollouts.
OmniVGGT~\citep{omnivggt} encodes RGB frames paired with their corresponding depth maps into geometry tokens $E_t$. We use cross-attention for condition injection, as in RDT~\citep{rdt}. With layers numbered from 1, we inject these tokens after backbone layers $\mathcal W=\{2,6,10,14,18,22,26\}$. At each injection layer, the hidden states provide cross-attention queries and $E_t$ provides keys and values. The attention output is scaled by a learnable coefficient and added to the hidden states.

Following the cross-layer aggregation principle of Attention Residuals~\citep{attnres}, a token-wise router combines the incoming hidden states with the outputs of all preceding injection layers before layers $\mathcal R=\{4,8,12,16,20,24,28\}$. Let $H_j$ be the output of backbone block $j$ before depth injection, $M_j$ its output after injection, and $X_r$ the input to block $r$ before routing. For $j\in\mathcal W$ and $r\in\mathcal R$, the two operations are
\begin{equation}
M_j=H_j+\gamma_j\,\CA_j(H_j,E_t),
\qquad
\widetilde X_r=\alpha_{r,0}\odot X_r
+\sum_{\substack{j\in\mathcal W\\j<r}}\alpha_{r,j}\odot M_j.
\label{eq:depthfusion}
\end{equation}
Here $\CA_j$ denotes cross-attention, $\gamma_j$ is a learnable residual scale, and $\odot$ applies a separate weight at each token position. The weights $\alpha_{r,\cdot}$ are computed from the candidate features and normalized across sources with a softmax. The model thus learns how much geometry to incorporate and how to combine features from different fusion layers. Small initial residual scales and a preference for the incoming state keep the new depth pathway close to the pretrained computation at initialization.

\begin{figure}[t]
\centering
\includegraphics[width=.90\linewidth]{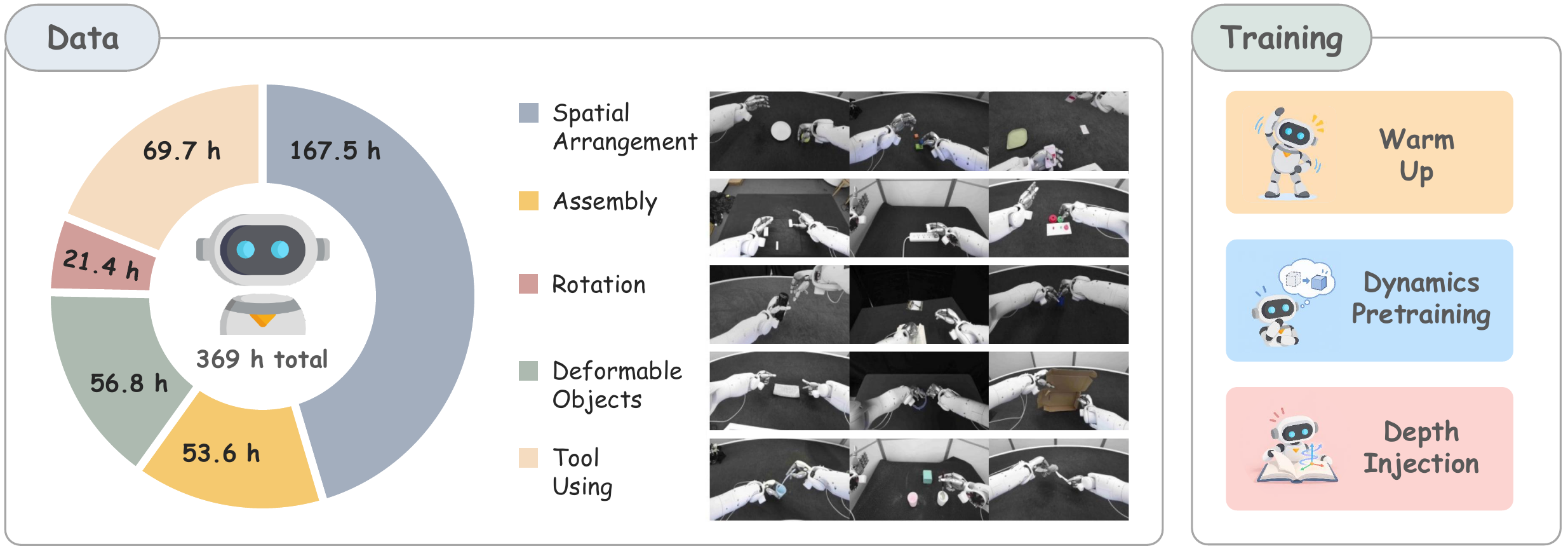}
\caption{\textbf{Real-robot data and training.} Recording hours and example frames for the five task categories. The final stage uses 100 hours with depth estimated offline by StereoAnyVideo.}
\label{fig:recipe}
\end{figure}

\paragraph{Action-conditioned world queries.}
A set of $N$ learnable world queries attends to hidden states from backbone layers 12, 20, and 28. Before action conditioning, these queries encode the current observations and hand-state history.

For future prediction, we use $K=4$ slots with offsets $\delta_k$, sampling one offset from each four-frame bin across a 16-frame future window. We express the demonstrated control at each offset relative to the current robot state and encode it with its temporal offset. The query adapter attends to these action tokens to produce $Z_t^{\mathrm w}$. The demonstrated controls disambiguate which future evolution should be modeled. The hand-flow and visual-dynamics objectives described below train $Z_t^{\mathrm w}$ to encode how the articulated hand and observed scene evolve under these controls. Thus, actions serve as conditioning inputs, while hand motion and visual changes provide the prediction targets. Both prediction experts receive $Z_t^{\mathrm w}$ and the same action tokens, conditioning their predictions on the controls associated with the target future states.

\paragraph{Predicting hand motion and visual changes.}
The Hand Flow Expert predicts displacements of corresponding hand points, while the Visual Dynamics Expert predicts changes in image features. At offset $\delta_k$, their targets are
\begin{equation}
F_{t,k}=P_{t+\delta_k}-P_t,
\qquad
\Delta V_{t,k}=\Phi(I_{t+\delta_k})-\Phi(I_t).
\label{eq:targets}
\end{equation}
Here, $P_t \in \mathbb{R}^{M\times 3}$ contains $M$ persistent hand anchors attached to different URDF links in the robot base frame, and $\Phi$ extracts and pools frozen DINOv3~\citep{dinov3} visual features from RGB. Thus, $\Delta V_{t,k}$ is a DINOv3 feature residual. Unlike a global hand or end-effector displacement, $F_{t,k}$ captures point-wise motion across the articulated hand surface, requiring the representation to encode how different hand links evolve under the demonstrated controls. This supervision also covers occluded fingers, while the visual residual captures the accompanying changes in the observed scene. Both targets are obtained from recorded robot states and RGB frames. Depth supplies current scene geometry and is not required at future timestamps.

Both experts predict noise added to their targets, with an additional Smooth L1 loss on the recovered clean targets. We apply \emph{future-horizon dropout} by sampling a prefix length $h\in\{1,\ldots,K\}$, exposing the first $h$ action tokens, and evaluating both prediction losses at the same offsets. The joint training objective is
\begin{equation}
\mathcal L_{\mathrm{world}}
= \mathcal L_{\mathrm{hand}}
+ \lambda_{\mathrm v}\mathcal L_{\mathrm{visual}}
+ \lambda_{\mathrm c}\mathcal L_{\mathrm{context}}.
\label{eq:worldloss}
\end{equation}
The auxiliary context loss reconstructs current hand, robot-state, scene, and instruction features from the queries before action conditioning. Both prediction losses backpropagate through $Z_t^{\mathrm w}$ and the backbone, training the query features that supervise the policy in Section~\ref{sec:policy}.

\paragraph{Data and training.}
For the real-robot setting, we organize approximately 370 hours of demonstrations across 60 tasks into five categories: spatial arrangement, assembly and contact, rotation and articulation, deformable objects, and tool use (Figure~\ref{fig:recipe}). These tasks cover changes in object placement, precise alignment, joint rotation, object shape, and sustained contact, providing varied examples of hand motion and scene changes for world-model training. We estimate depth offline with StereoAnyVideo for a 100-hour subset covering spatial arrangement, assembly and contact, and rotation and articulation, matching the three categories evaluated on the real robot.

Training proceeds in three stages: \emph{warmup}, \emph{dynamics pretraining}, and \emph{depth injection}. During warmup, we jointly optimize the pretrained backbone and the newly introduced prediction modules, allowing the backbone representations to adapt to the hand-motion and visual prediction objectives. We then perform dynamics pretraining on the demonstration corpus, where the model learns action-conditioned interaction dynamics from RGB observations, language instructions, and histories of hand and robot states. These first two stages are conducted without depth input.

In the final depth-injection stage, we introduce depth conditioning and continue training on the 100-hour subset. Geometry features are extracted by a frozen encoder and injected into the backbone through trainable fusion modules, while the backbone and prediction modules remain jointly optimized. The hand-motion and visual prediction objectives remain unchanged, allowing depth conditioning to refine the learned dynamics representations under consistent predictive supervision.

\subsection{Injecting Spatial Foresight into Policy Latents}
\label{sec:policy}

After world-model training, we freeze its parameters and use the action-conditioned queries $Z_t^{\mathrm w}$ to supervise policy queries $Z_t^\pi$, as shown by the query-alignment connection in Figure~\ref{fig:dexforesight}. Computing these targets requires the world-model backbone and query adapter, without executing either prediction expert. We jointly optimize query alignment and action prediction to train the policy.

\paragraph{Reading queries and conditioning actions.}
We add $N$ learnable policy queries that attend to selected hidden layers of the policy VLM. A learned projection maps their outputs into the world-query feature space for alignment. These queries receive only the policy's current inputs.

Let $C_t$ denote the action expert's original context. A learned projection $W_{\mathrm{act}}$ maps the policy queries to the action-context dimension, after which we concatenate the tokens: $\widetilde C_t=[C_t;Z_t^\pi W_{\mathrm{act}}]$. The action expert uses $\widetilde C_t$ to generate $A_t$. This gives it access to the query features alongside its original context. Alignment gradients update the query reader and the trainable VLM parameters; action gradients also train how the expert uses the added tokens.

\paragraph{Aligning query features.}
We match corresponding query features and their pairwise similarities. Let $\bar Z^\pi$ and $\bar Z^{\mathrm w}$ contain all $N$ policy and world queries after layer normalization and unit-length normalization; time subscripts are omitted below. The teacher features are detached. Define $K=\bar Z\bar Z^\top$ as the pairwise cosine-similarity matrix. The policy objective is
\begin{equation}
\begin{gathered}
\mathcal L_{\mathrm{token}}
=\frac{1}{N}\sum_{i=1}^{N}
\bigl(1-\langle\bar z_i^\pi,\bar z_i^{\mathrm w}\rangle\bigr),
\qquad
\mathcal L_{\mathrm{relation}}
=\frac{\lVert K^\pi-K^{\mathrm w}\rVert_F^2}{N^2}, \\[4pt]
\mathcal L_{\mathrm{policy}}
=\mathcal L_{\mathrm{action}}
+\lambda_{\mathrm{align}}\bigl(
\mathcal L_{\mathrm{token}}+\beta\mathcal L_{\mathrm{relation}}\bigr).
\end{gathered}
\label{eq:guidance}
\end{equation}
The token loss matches each policy query to its teacher counterpart, and the relation loss preserves the teacher's pairwise query similarities. We retain the policy's original action objective. Since the policy does not observe $U_t$, query alignment is an auxiliary learning target, not an exact reconstruction of the teacher's action-conditioned predictions. Demonstrated controls specify the dynamics used for supervision; the policy learns from the resulting control-dependent visual and hand-motion variation while retaining its standard behavior-cloning objective.

\paragraph{Deployment.}
At inference, the policy reads current RGB images, language, and robot state, computes its queries, and supplies the augmented context to the action expert. It computes this context once per observation and reuses it during action sampling. The world model, depth encoder, hand-geometry encoder, and both future experts are removed. The deployed policy retains the query reader and context projection, with its original action representation and sampling procedure.

\section{Experiments}
\label{sec:experiments}

Our experiments answer three research questions:
\begin{itemize}
    \setlength{\itemsep}{0.15em}
    \setlength{\parsep}{0pt}
    \item \textbf{RQ1 (baselines).} Can DexForesight, which decodes no future at deployment, outperform vision--language--action (VLA) baselines and world action models that explicitly learn future prediction, both in simulation and on real dexterous hardware?
    \item \textbf{RQ2 (spatial foresight and transfer).} Is the predictive information learned by the world model retained and accessible in the deployed policy representation, and is transferring these features more effective than direct prediction?
    \item \textbf{RQ3 (world-model ablation).} How do offline depth, cross-layer depth fusion, and the 3D hand-motion target contribute to policy performance?
\end{itemize}
\subsection{Evaluation Setup}
\label{sec:eval_setup}

\begin{figure}[t]
\centering
\includegraphics[width=.88\linewidth]{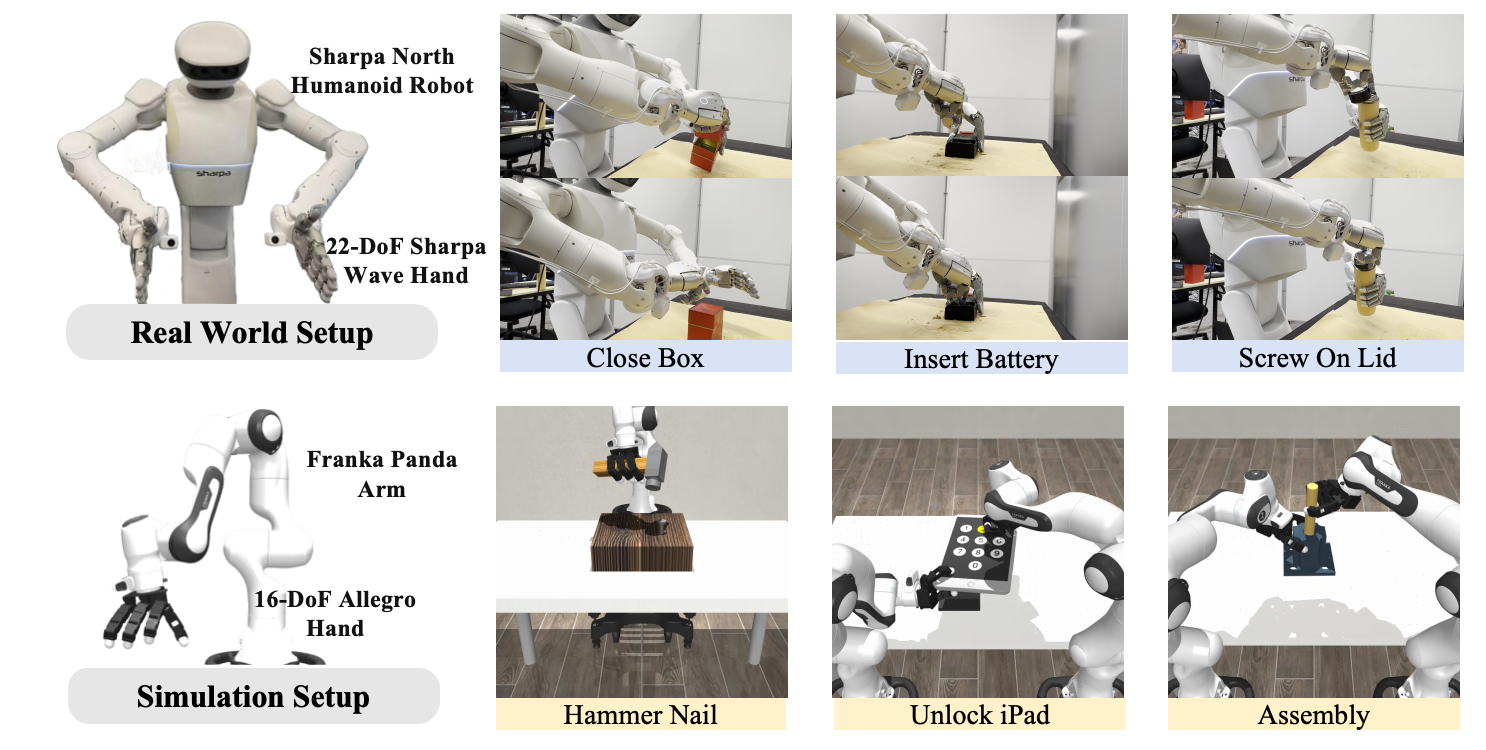}
\caption{\textbf{Experimental setup and tasks.} Top: the Sharpa North robot with 22-DoF Sharpa Wave hands and the three real-robot tasks. Bottom: the Franka Panda with a 16-DoF Allegro hand and example simulation tasks.}
\label{fig:experiment_setup}
\end{figure}

Figure~\ref{fig:experiment_setup} shows the robot platforms and example tasks.

\paragraph{Simulation.}
We evaluate 11 DexJoCo tasks~\citep{dexjoco} spanning single-arm and bimanual manipulation. In simulation, DexForesight augments GR00T N1.7~\citep{grootn1} and is compared with GR00T N1.7, $\pi_{0.5}$~\citep{pi05}, Fast-WAM~\citep{yuan2026fast}, and Cosmos3~\citep{cosmos}. The spatial world model is fine-tuned on DexJoCo training demonstrations. Depth is rendered directly from MuJoCo rollouts, without stereo depth estimation. We report task success rates.


\paragraph{Real robot.}
On the real robot, we instantiate the same DexForesight formulation on QwenVLA~\citep{qwenvla}, which is pretrained on the approximately 370-hour corpus. We select one challenging task from each depth-stage category: Close Box (spatial arrangement), Insert Battery (assembly and contact), and Screw On Lid (rotation and articulation), abbreviated as Close, Insert, and Screw. All trials use held-out object positions; Screw also varies cup color. Each method is fine-tuned for 30 epochs and evaluated over 20 trials per task. WLA augments QwenVLA with Wan, and Fast-WAM, GR00T N1.7, and Cosmos3 follow their official fine-tuning procedures.


\subsection{RQ1: Baseline Comparison}
\label{sec:rq1}

\emph{Can DexForesight retain useful predictive structure without future decoding at deployment and outperform VLA and world-action baselines?}

\paragraph{Simulation.}
Table~\ref{tab:sim_baselines} compares DexForesight with two VLA policies ($\pi_{0.5}$ and GR00T N1.7) and two world action models (Fast-WAM and Cosmos3) on the 11 DexJoCo tasks. DexForesight reaches 60.5\% overall success: 9.6 percentage points above the strongest baseline, Fast-WAM, and 11.1 points above its own backbone, GR00T N1.7. It also achieves the highest success in each task group, at 76.3\% single-arm and 41.6\% bimanual. At deployment, DexForesight generates actions from policy queries and runs neither the spatial world model nor future decoding.

\begin{table}[t]
\centering
\small
\caption{\textbf{Simulation success rates (\%).} 100 trials per task.}
\label{tab:sim_baselines}
\begin{tabular}{lrrr}
\toprule
Method & Single-arm & Bimanual & All \\
\midrule
$\pi_{0.5}$ & 53.0 & 13.6 & 35.1 \\
GR00T N1.7 & 61.5 & 34.8 & 49.4 \\
Fast-WAM & 60.2 & 39.8 & 50.9 \\
Cosmos3 & 59.5 & 37.4 & 49.5 \\
DexForesight (based on GR00T N1.7) & \textbf{76.3} & \textbf{41.6} & \textbf{60.5} \\
\bottomrule
\end{tabular}
\end{table}

\paragraph{Real robot.}
On physical hardware, WLA enables its World Expert, Cosmos3 jointly generates future video and actions, and Fast-WAM skips future generation at inference. DexForesight generates actions directly from its aligned policy queries. WLA~\citep{yang2026wla} and Fast-WAM~\citep{yuan2026fast} are external comparisons to alternative train--deploy strategies; architectural and training differences preclude treating either as a one-variable causal control. Table~\ref{tab:real_baselines} reports held-out-position results. DexForesight reaches 58.3\% overall success, versus 50.0\% for WLA, 48.3\% for GR00T N1.7, 30.0\% for Cosmos3, and 10.0\% for Fast-WAM. Failures reflect the intended spatial challenges: finger occlusion in Screw, partial object occlusion in Insert, and lid--box geometry in Close. Under identical hardware and inference settings, policy inference takes 227.4\,ms for QwenVLA and 230.4\,ms for DexForesight, an increase of only 3.0\,ms (1.3\%). Cosmos3 takes 930\,ms to generate a 16-frame future video, with this measurement covering its video-generation stage. Section~\ref{sec:rq2} provides controlled evidence for the transferred representation.

\begin{table}[t]
\centering
\small
\caption{\textbf{Real-robot success rates (\%).} 20 trials per task. Screw / Close / Insert abbreviate Screw On Lid / Close Box / Insert Battery.}
\label{tab:real_baselines}
\begin{tabular}{lrrrr}
\toprule
Method & Screw & Close & Insert & Overall \\
\midrule
GR00T N1.7 & 40.0 & 35.0 & 70.0 & 48.3 \\
QwenVLA & 40.0 & 15.0 & 35.0 & 30.0 \\
WLA & 50.0 & 30.0 & 70.0 & 50.0 \\
Fast-WAM & 15.0 & 5.0 & 10.0 & 10.0 \\
Cosmos3 & 25.0 & 15.0 & 50.0 & 30.0 \\
DexForesight & \textbf{50.0} & \textbf{45.0} & \textbf{80.0} & \textbf{58.3} \\
\bottomrule
\end{tabular}
\end{table}

\subsection{RQ2: Spatial Foresight and Representation Transfer}
\label{sec:rq2}

\emph{Is the predictive information learned by the world model retained and accessible in the deployed policy representation, and is transferring it more effective than directly attaching prediction heads to the policy?}

\paragraph{Spatial foresight probe.}
We test whether predictive information remains accessible in deployed policy queries rather than only improving task success. We freeze real-robot checkpoints and train a linear ridge probe from policy features to future 3D hand-point displacements at 0.13, 0.27, and 0.54\,s; feature extraction uses neither the world model nor its prediction experts. DexForesight and QwenVLA action-only share architecture and feature extraction. The QwenVLA action-only model uses the same checkpoint as the QwenVLA baseline in Table~\ref{tab:real_baselines}. DexForesight is initialized from this pretrained checkpoint and further trained with query alignment and action prediction. GR00T N1.7 is a cross-architecture reference using its native final-layer visual representation. Evaluation uses episode-disjoint trajectories and 95\% episode-bootstrap confidence intervals.

\begin{table}[!ht]
\centering
\small
\caption{\textbf{Future hand-motion information in frozen deployed latents.} Linear-probe 3D displacement error in mm ($\downarrow$); brackets are 95\% episode-bootstrap CIs. Features are extracted from the policy alone.}
\label{tab:spatial_probe}
\begin{tabular}{lccc}
\toprule
Method & 0.13\,s & 0.27\,s & 0.54\,s \\
\midrule
DexForesight &
\textbf{3.92 [3.47, 4.13]} &
\textbf{6.33 [6.01, 7.07]} &
\textbf{10.01 [9.52, 11.01]} \\
QwenVLA action-only &
4.12 [3.83, 4.47] &
7.70 [7.32, 8.09] &
13.59 [13.01, 14.21] \\
GR00T N1.7 &
4.39 [4.12, 4.67] &
7.99 [7.66, 8.30] &
13.74 [13.14, 14.42] \\
Zero-motion prior &
5.08 [4.59, 5.61] &
9.79 [8.93, 10.62] &
18.48 [16.99, 19.83] \\
\bottomrule
\end{tabular}
\end{table}

DexForesight yields lower probing error than the architecture-matched QwenVLA policy at every horizon, with the gap increasing over time. At 0.54\,s, the error falls from 13.59 to 10.01\,mm, a 26.3\% reduction. Thus, predictive hand-motion information learned by the world model remains more linearly accessible in the deployed DexForesight policy queries even after the world model and future decoders are removed.

\paragraph{Transfer versus direct prediction.}
We next ask whether directly attaching predictive objectives to the policy can provide the same benefit. The upper block of Table~\ref{tab:ablations} compares DexForesight with four direct additions to GR00T: a DINO future-residual head, a hand-flow head, both heads together, and depth with AttnRes. The hand-flow head is the strongest direct-prediction control at 53.1\% overall, 3.7 points above GR00T. DexForesight nevertheless reaches 60.5\%, exceeding this control by 7.4 points, while combining both prediction heads reaches only 49.5\%. Together with the latent probe, these results suggest that policy latents can retain predictive structure useful for control without future decoding. In this setting, query alignment is more effective than directly supervising visual-change or hand-motion prediction.

We further examine the teacher target. Pre-action queries encode scene and hand context without explicit conditioning on the demonstrated controls, yielding 52.7\% versus 49.4\% for the GR00T N1.7 backbone. Their weaker performance is consistent with a gap between observation-level context and the specific manipulation behavior being learned. Conditioning on the demonstrated actions provides control-dependent predictive features, raising success to 60.5\%. Replacing hand-motion supervision with a future-action target reaches only 48.4\%, suggesting that action prediction alone does not provide an equally effective transfer target. The teacher is trained, including through future-horizon dropout over exposed action prefixes, to model how actions affect future evolution. Shuffling the conditioning actions reduces success to 41.5\%, below the GR00T N1.7 backbone. This is consistent with mismatched controls disrupting the correspondence between the demonstrated behavior and the teacher's predictive spatial guidance.

\begin{table}[!h]
\centering
\small
\setlength{\tabcolsep}{6pt}
\renewcommand{\arraystretch}{1.08}
\caption{\textbf{Simulation ablations (\%).} Single-arm, bimanual, and overall success rates across 11 DexJoCo tasks.}
\label{tab:ablations}
\begin{tabular}{lrrr}
\toprule
Variant & Single-arm & Bimanual & All \\
\midrule
DexForesight & \textbf{76.3} & \textbf{41.6} & \textbf{60.5} \\
\midrule
\multicolumn{4}{l}{\emph{Direct additions to GR00T}} \\
DINO future-residual head & 63.5 & 23.8 & 45.5 \\
Hand-flow head & 66.0 & 37.6 & 53.1 \\
DINO and hand-flow heads & 64.7 & 31.4 & 49.5 \\
Depth and AttnRes & 59.2 & 33.2 & 47.4 \\
\midrule
\multicolumn{4}{l}{\emph{DexForesight query-target ablations}} \\
Pre-action world queries & 64.2 & 39.0 & 52.7 \\
Shuffled action conditioning & 58.5 & 21.0 & 41.5 \\
\midrule
\multicolumn{4}{l}{\emph{DexForesight world-model ablations}} \\
No offline depth & 63.2 & 33.6 & 49.7 \\
Context depth tokens & 52.3 & 29.6 & 42.0 \\
Future-action target & 62.7 & 31.2 & 48.4 \\
\bottomrule
\end{tabular}
\end{table}
\FloatBarrier

\subsection{RQ3: World-Model Components}
\label{sec:rq3}

\emph{How do offline depth, cross-layer depth fusion, and the 3D hand-motion target contribute to policy performance?}

The final block of Table~\ref{tab:ablations} removes or replaces one component at a time. Without offline depth, success drops by 10.8 points in simulation and 15.0 points on the real robot, from 58.3\% to 43.3\% (35.0\% Screw, 25.0\% Close, and 70.0\% Insert), while deployed inputs remain unchanged. Replacing cross-layer fusion with context depth tokens in the RynnBrain-initialized world model yields 42.0\%, indicating that the teacher's depth-fusion mechanism substantially affects downstream policy performance. We hypothesize that concatenating depth tokens perturbs the pretrained multimodal features. Replacing hand-point displacement with future-action targets yields 48.4\%, supporting the use of kinematic hand-motion supervision, which also covers occluded hand surfaces. Together, these results support transferring predictive representations learned with training-only depth, hand geometry, and future-state supervision into the policy.


\section{Conclusion}
\label{sec:conclusion}

DexForesight grounds dexterous policy latents in predictive hand--scene dynamics learned from offline depth, kinematic hand geometry, and demonstrated controls. Latent alignment transfers these dynamics into a policy that uses only RGB, language, and robot state at deployment. Across simulation and real-robot tasks, the transferred representation improves success and preserves future hand-motion information after the world model is removed.

\end{document}